# Bio-inspired Tensegrity Soft Modular Robots

D. Zappetti, S. Mintchev, J. Shintake, and D. Floreano

Laboratory of Intelligent Systems at Ecole Polytechnique Fédérale de Lausanne (EPFL),
CH1015 Lausanne, Switzerland
`davide.zappetti@epfl.ch`

**Abstract.** In this paper, we introduce a design principle to develop novel soft modular robots based on tensegrity structures and inspired by the cytoskeleton of living cells. We describe a novel strategy to realize tensegrity structures using planar manufacturing techniques, such as 3D printing. We use this strategy to develop icosahedron tensegrity structures with programmable variable stiffness that can deform in a three-dimensional space. We also describe a tendon-driven contraction mechanism to actively control the deformation of the tensegrity modules. Finally, we validate the approach in a modular locomotory worm as a proof of concept.

## 1   Introduction

The quest for reconfigurable modular robots dates back to von Neumann's early speculations that one would need a dozen types of simple modules fabricated in millions and floating in an agitated medium [1] to Fukuda's cellular robots composed of self-contained mobile units that could assemble to carry out diverse tasks [2]. Over the past 30 years, this challenge has been tackled by the growing fields of modular robotics [3] and swarm robotics. While there is not enough space here to review those two fields, it is important to point out that modular and swarm systems fall in two categories: systems made of complex units that can individually move or perform some tasks and systems made of simple units that can move or perform tasks only when they are assembled with other units. In this paper we are concerned with the latter type of systems where morphofunctional properties emerge from the assembly of multiple units, as in biological multi-cellular organisms. Since most modular artificial systems are made of rigid components, they display some scalability challenges [4]: they may require precise motion for docking of terminal units, the morphological space may be limited to simple geometries, and the resulting structure may be relatively rigid for robust and safe interaction with the environment.

On the other hand, biological multi-cellular systems are made of cells that display different degrees of stiffness: very soft, such as fat cells, very stiff, such as bone cells, to directional stiffness, such as hair cells, and variable stiffness, such as muscle cells, for example. A soft modular robot made of programmable-stiffness modules could display physical compliance providing safer and better interaction with the environment and span a larger morpho-functional space, including curved bodies, soft bodies, such as worms, and complex bodies, such as vertebrates.

Recently, some authors have described examples of soft modular robots and most of them make use of pneumatic actuation and inflatable modules [5-7]. J. Y. Lee et al. [8] proposed a Soft Robotic Blocks kit (SoBL) composed of three basic types of pneumatically actuated soft modules. Each type of module implements a single type of motion (i.e. translation, bending or twisting) and can be assembled in branch-like structures to achieve more complex motion patterns. However, pneumatic actuation requires several pumps and complex network of pipes that make the system complex and bulky for a high number of independent controlled modules. Non-pneumatic approaches have also been used. Yim et al [9] described a chain of soft cubes that could return to a previously designed 3-D shape after being stretched. Wang et al. [10] describe a variable stiffness actuated hinge that can be used to form modules capable of being assembled in various different deployable structures. B. Jenett at al. [11] describe the use of two-dimensional compliant modules assembled in three-dimensional structures. The stiffness and density of the assembly can be programmed locally based on the number and the orientation of the modules. Some of us have recently described a soft multi-cellular robot [12] where each cell is a ring made of PDMS at desired stiffness value with embedded magnets. The cells could connect in chains and, by automatically selecting the appropriate cell stiffness and the position of the magnetic contact points, they could approximate a large variety of desired curved morphologies by magnetic repulsion or attraction of the magnets in adjacent cells. The cells have also been functionalized by adding an internal network of nitinol fibers connecting opposite points of the ring that could contract the cell in desired directions, akin to muscle cells [13]. However, the system was constrained to two dimensions.

Here we propose a design principle for three-dimensional soft modules based on tensegrity structures inspired by the cytoskeleton of biological cells. Tensegrity structures are lightweight, can undergo large deformations generated by internal or external forces, and can resist large compressive forces. We describe a method for designing and assembling three-dimensional modules with planar manufacturing technologies. Moreover, the stiffness of the cells can be programmed by changing some parameters of the manufacturing process. We also describe a method to add contraction movement by means of tendon-driven actuation and validate the approach in a proof-of-concept crawling, multi-modular worm.

The paper is organized as follows: in section II, the tensegrity approach is introduced. In section III the main tensegrity structure module is selected, and materials, manufacturing tools and processes are described. The tendon-driven contractive actuation that can be added to the module is also presented. In section IV a proof of concept soft modular crawling robot is presented. Finally, in section V we discuss future work.

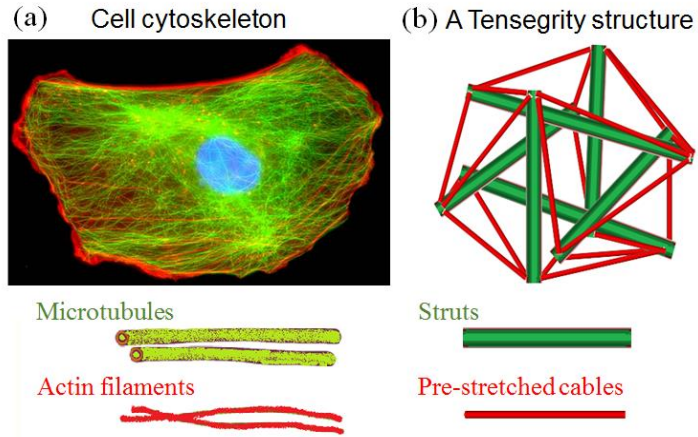

**Fig. 1.** (a) Image of a cell obtained with fluorescence microscopy. In blue is the nucleus of the cell and at the bottom a sketch of the two main constituents of the cytoskeleton: microtubules and actin filaments [14]. (b) Example of a tensegrity structures: the icosahedron tensegrity structure composed of six struts and 24 cables.

## 2    Tensegrity in biological and robotic systems

In order to develop soft modules with programmable stiffness for our modular robot, we took inspiration by the mechanical structure of multicellular organisms cells. These organisms are heterogeneous systems made of cells with very diverse functions and mechanical properties. However, every cell has a cytoskeleton, which is an extremely complex network made of two types of interconnected fibers [14]: rigid microtubules and bendable actin filaments (Fig. 1a). This architecture, which is responsible for the shape, stiffness, and strength of the cell can be formally described as a tensegrity structure [15].

The term "tensegrity" has been coined by architect R. Buckminster Fuller to describe a structure that maintains its mechanical integrity throughout the pre-stretching of some elements constantly in tension (called "cables" in red in Fig. 1b) connected in a network with other elements constantly under compression (called "struts" in green in Fig. 1b) [15]. In the cytoskeleton, the microtubules and the actin filaments function as struts cables, respectively. The cytoskeleton has different values of stiffness (e.g. resistance to external loads) according to the level of pre-stretch of its actin filaments (i.e. the higher the pre-stretch, the higher the stiffness of the cytoskeleton) [15]. Similarly, the stiffness of a tensegrity structure with a given network configuration depends on the pre-stretch of its cables. If the pre-stretch is properly tuned, the tensegrity structure has low stiffness in all directions (e.g., as in elastomers and living matter), and display a behavior akin to soft matter [15].

In addition to describing the cytoskeleton of the cells, the tensegrity model has been used to describe several structures at many scales of the life [16]. An example at the macroscopic level is the skeletal system of the human body, where bones, tendons and muscles are elements continuously under tension or compression to keep the body in its mechanical integrity [16] [17]; an example at the nanoscale is some proteins and basic molecules that maintain integrity with tensegrity structure [17]. The scalability of the tensegrity model is therefore an important asset for modular robot design because it could be employed at different levels, although manufacturing methods may vary according to the chosen scale.

The use of tensegrity structures to develop biologically inspired robots has already been suggested by Haller et al. [18], while Rieffel et al. suggested the use of tensegrities to achieve "mechanical intelligence" [19]. Yet another example of a tensegrity robot is SUPERball developed at NASA [20], which is able to roll, even on rough terrains, with an optimized control of the six embedded actuators. However, to the best of the author's knowledge, there have not yet been proposals of using tensegrity structures as a design principle for simple modules of a modular, functional robot.

## 3     Module Design and Manufacturing

In this paper we aim at studying robotic modules in the scale of few centimeters because they can be manufactured using affordable and simple manufacturing methods, incorporate off-the-shelf electronic components, and eventually be assembled into functional modular robots of a size comparable to typical household or inspection robots. In this section we describe the selection of the main tensegrity structure module, the choice of materials, and the manufacturing method.

### 3.1     Tensegrity Structure Selection

Different tensegrity structures can be obtained according to the number of struts and cables, the network configuration (e.g. struts and cables positions in the three-dimensional space), the cables' stiffness and how much they are pre-stretched [21]. In this study, for sake of simplicity, we assume that all the cables of the tensegrity structure have the same pre-stretch value.

Three main criteria have been applied to select the main tensegrity structure for a modular tensegrity robot. The first criterion is that the main soft module should be able to deform (e.g. stretch or contract) in all directions to augment the morphological diversity of the assembly and comply with objects and surfaces in a three-dimensional space. The second criterion is that the tensegrity structure should involve the smallest possible number of struts and cables to ease the manufacturing and assembling. The third criterion is to favor network configurations whose inner volume is not crossed by any cable or strut in order to place and protect a useful payload (e.g. actuator, microcontroller, energy storage, sensor, depending on the cell type).

To assess the first criterion, we had to consider that tensegrity structures can display different values of stiffness along different directions. However, the more the structure is symmetric, the more it will exhibit similar mechanical properties along different directions [21]. The tensegrity structures that use the most symmetric networks are those with an almost spherical shape [22]. Among these, to assess the second criterion, we selected the tensegrity structure that has the smaller number of struts and cables: the icosahedron tensegrity (Fig. 2) [20] [22]. It is composed of 6 identical struts and 24 cables of equal length. The cables are organized in 8 equilateral triangles interconnected by 12 vertices and distributed in 4 parallel opposite pairs (in Figure 2b the pairs of triangles are marked with four different colors). Furthermore, the icosahedron tensegrity structure has an inner cubic volume that is not crossed by any cable or strut (Fig. 2a), satisfying the third criterion.

When the icosahedron structure is compressed along a direction orthogonal to any of the four triangle pairs (Fig 2c), it displays maximum deformability and can be collapsed to a flat configuration (Fig. 2d). These four directions, which we name "collapsibility directions", allow the structure to deform in three-dimensional space.

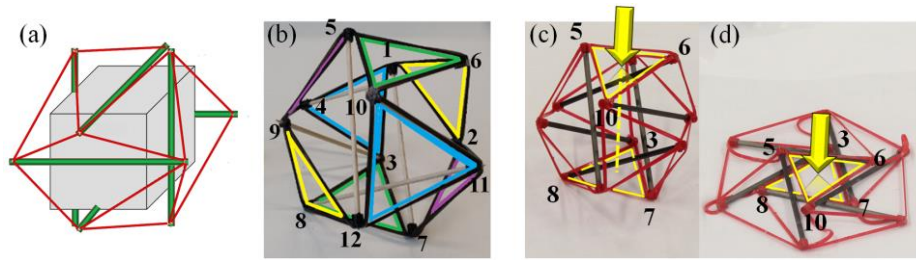

**Fig. 2.** (a) Icosahedron tensegrity structure with a grey cube at the geometrical center to better display the inner cubic volume not crossed by any cable or strut. (b) Icosahedron tensegrity prototype with the 4 couples of parallel equilateral triangular faces marked with 4 different colors. (c) An external load is applied along the orthogonal direction to the two parallel triangular faces (a "collapsibility direction"). (d) The collapsed structure.

### 3.2 Design and Manufacturing

Instead of manufacturing every single component separately, here we propose to manufacture all cables as a single flat network that can be folded into a three-dimensional structure and subsequently filled with struts. The cable network can be rapidly manufactured using inexpensive 3D printers.

Two different materials are used for cables and struts. The cables require an elastic printable material that can withstand a wide range of pre-stretch and deformations of the module without losing its elastic properties; the struts instead require a stiffer material that can withstand compressive forces without buckling [21]. For cables, we use NinjaFlex, an elastic material compatible with commercially available fused deposition modeling 3D printers that can withstand 65% of elongation at yield. For struts, we use pultruded carbon rods (that are commercially available in different diameters and length) with a longitudinal Young's modulus of approximately 90 GPa, which is sufficient to withstand compression and buckling.

To design the flat cable's network, a 3D CAD model of the tensegrity module is realized (Fig. 3a). The model is composed of the 8 equilateral triangles of the icosahedron tensegrity assembled with joints at the 12 vertices. The unfolded flat network is obtained by eliminating the 6 struts and disconnecting the cables at two vertices (see Fig

3b), and then rotating the triangles around the joints until obtaining a flat network's configuration (see Fig. 3c-d). The nodes of the network are marked according to the corresponding vertex of the 3D tensegrity model. The two vertices that were disconnected in software to unfold the network will be overlapped to close it back during the physical assembly.

To assemble the rigid struts in the elastic network, cylindrical housings are designed at all the nodes (see Fig. 3d-e). The housings are 4 mm high and have an inner diameter equal to the one of the carbon rods in order to ease the insertion. Thanks to this type of design, no additional connection elements or adhesives are required to connect the struts, which will be kept in place and secured by the pre-stretch of the cables after the physical assembly.

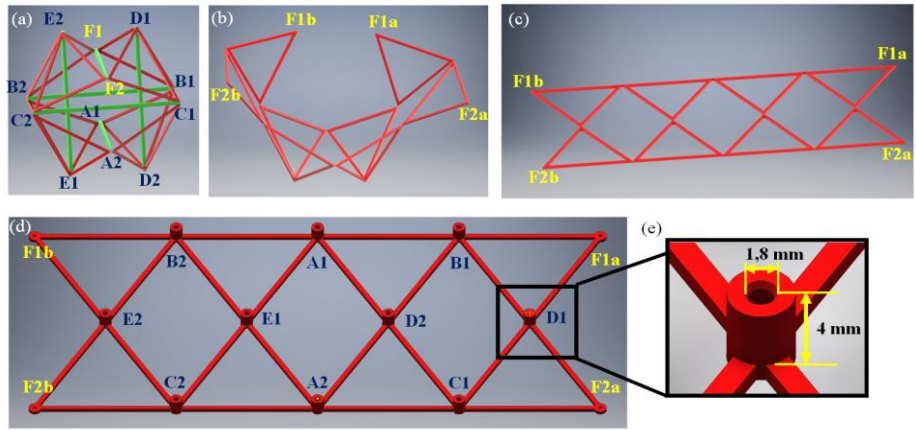

**Fig. 3.** (a) 3D CAD model of a tensegrity icosahedron. In yellow are marked the two vertices disconnected to unfold the cable's network. (b-c) The unfolding sequence obtained rotating the triangles around the joints in the vertices, in yellow are marked the 4 nodes that will be overlapped to generate vertices F1 and F2. (d) Flat cable's network with housings in the vertices to insert the carbon rods. (e) Detailed view of one of the housings.

The cable network is 3D printed with a Desktop 3D printer LulzBot TAZ 5. After 3D printing, the carbon rods are assembled following the illustration in Fig. 3d: the two vertices of a rod are inserted in the two housings with the same letter starting from the housing "A" (e.g., the ends of the first strut will be inserted in A1 and A2 housings) and following the alphabetical order until the sixth and final strut "F" which has to be inserted in four housings (one side of the carbon rod is inserted in the two F1a-b housings and the other in the two F2a-b).

The icosahedron tensegrity prototype in Fig. 2c is made of an elastic network of thickness and width of 1 mm and with cables long 4.75 cm each, has a height of 7.8 cm and can be approximated to a sphere of the same diameter, therefore occupying a volume of about 248 $cm^3$. The printing process requires approximately 30 minutes for the icosahedron elastic network with an infill of 100 % and 0.25 mm of vertical resolution. When collapsed along any of the four different directions, the structure reduces its volume by 84 % to about 40 $cm^3$ (Fig. 2d).

The stiffness of the tensegrity modules mainly depends on the stiffness of the cable networks that deforms when the module is compressed. Hence, by modifying the cable's stiffness, it is possible to tune the elastic behavior of the entire module. This can be achieved by changing some design parameters, such as the thickness, width, material of the cables or their pre-stretch, which is defined by the ratio between struts and cables lengths. All cables have the same length and, if shortened, they become more stretched during the assembly (when the struts length is kept constant) and result into a stiffer module. Different prototypes with different pre-stretches and thicknesses of the cables have been manufactured and tested. The pre-stretch ranges from less than 1 % up to 30 % (after which the manual assembly becomes difficult) and the thickness from 1mm to 3mm. Every module has been compressed 50% of its height and for every module a load versus compression curve has been obtained. The figure 4a shows increasing stiffness of the modules with increasing values of pre-stretch at a constant thickness of 1 mm, although there is no sensitive variation over 15%. Figure 4b shows higher stiffness with increasing thickness.

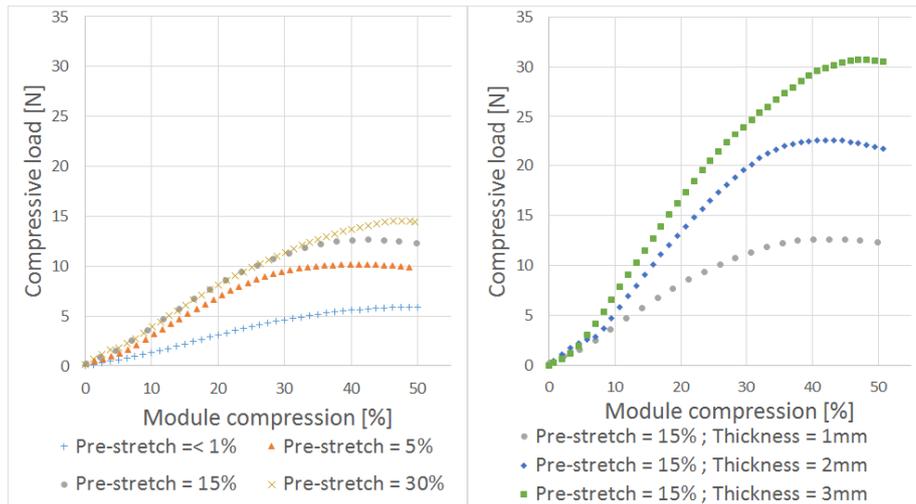

**Fig. 4.** (a) Compressive load versus module compression of manufactured modules with different pre-stretch and fixed 1mm thickness. The compression tests have been conducted with an Instron universal testing machine. (b) Modules with 15% pre-stretch and different thicknesses.

### 3.3 Connectivity

A connection mechanism is required to physically connect modules into a modular robot that can display some functionality. Connection points may also serve as a medium to transmit information and energy throughout the robot. Furthermore, the number and type of connection points in a module may affect the morphology and functionality of the robot. However, at this stage we use only a simple mechanical latching that allows us to assemble cells into a proof-of-concept functional robot.

The mechanical latching system connects vertices and faces of different modules. The system is integrated into the flat cable's network configuration and therefore manufactured with all the other cables during the 3D printing process (see Fig. 5a). The system consists of a pin and a hole at every vertex. The pin has a slightly larger diameter than the hole and when forced into it, it remains thanks to the friction between the lateral walls of pin and hole. An opposite pressure is required to detach them. The pin and hole can be inserted inside one another when the vertex is not connected to another vertex (see Fig 5b-c) or can be latched to another vertex with each pin inserted in the hole of the other vertex (see Fig. 5d). Two triangular faces of two modules can be connected latching their 6 vertices in 3 pairs (see Fig. 5e).

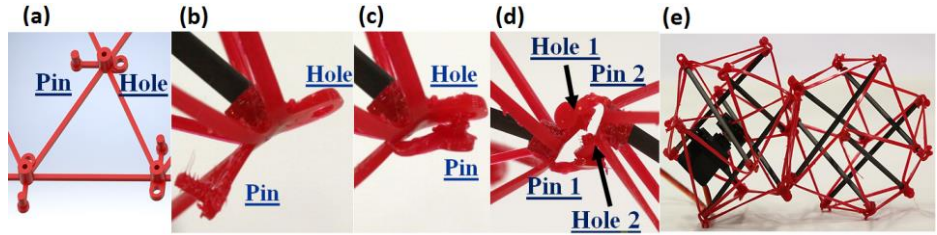

**Fig. 5.** (a) Integration of the latching system in the cable's network. (b) Pin and hole of a vertex disconnected. (c) Pin and hole of a vertex connected. (d) Pin and hole of a vertex connected to another vertex. (e) Two triangular faces connected through their 6 vertices.

### 3.4  Tendon-Driven Actuation

An actuation mechanism can be added to the main module in order to control its contraction along one of the four collapsibility directions. Although this type of 1DOF actuation does not allow a single module to locomote, just like single biological cells in multicellular systems, several actuated modules connected in series, in parallel or transversally could perform more complex movements and tasks, such as locomotion or manipulation.

In this paper, we propose to use a tendon-driven contractive system operated by a servo-motor. Although the servo-motor is a rigid component that can reduce the overall softness of the system [23] and the volume reduction in fully collapsed state, the chosen motor has a small volume (5 cm$^3$) compared to that of the module (248 cm$^3$). The servo is strategically placed in a modified strut with rectangular housing. This design avoids rigid connections between two struts and the rigid servomotor and preserve the tensegrity nature of the structure and its deformability [21].

The tendon-driven actuation contracts the icosahedron tensegrity by simultaneously pulling two opposite triangular faces towards the geometrical center of the structure along the collapsibility direction (see Fig 6a). The contractive mechanism comprises six tendons that connect each vertex of the selected triangular faces to a pulley that is activated by a servo-motor (see Fig 6b). The pulley is placed at the geometrical center of the icosahedron structure (see Fig 6b). When the servo is activated, the pulley starts to rotate wrapping the tendons, which in turn produce a contraction of the icosahedron (see Fig 6c).

The six-tendon design can be directly included in the design process of the module's elastic network. The six tendons have one end attached to the vertices of the triangular faces and the other end free (see Fig 6d-e).

The kinematic tests of the actuated module show that a compression of about 25 % (the negative pick at about 3.5 seconds in Fig. 6g) and a lateral diameter expansion of 9 % of the module height can be achieved. Improved design or different actuation technologies could further increase the contraction of the module, if required.

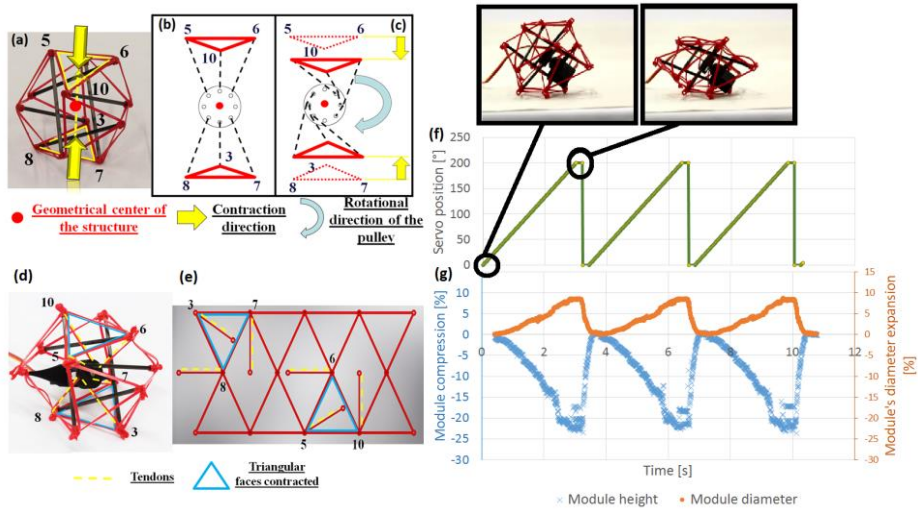

**Fig. 6.** (a) The arrows represent the contractive directions of the two triangular faces marked in yellow. While the red dot represents the geometrical center of the tensegrity structure toward which the two faces are contracted. (b) The six dashed lines represent the six tendons connecting the two triangular faces to the pulley placed at the geometrical center of the structure. (c) When the pulley is rotated, the six tendons are all pulled at the same time contracting the two triangular faces. (d) The assembled module with marked in dashed yellow lines the six tendons connecting the 6 vertices of the two opposite triangular faces (marked in blue) to the pulley in the geometrical center of the structure. (e) The unfolded network with the tendons. (f) Graph of the servo position driving signal generated by an Arduino Uno. (g) Corresponding graph of the module compression and lateral expansion recorder through motion capture system (i.e. Optitrack system recording at 240 Hz).

## 4    A Crawling Modular Robot

We exploited the specific kinematic of the actuated module (i.e. lateral expansion when compressed) to develop a simple crawling modular robot using peristaltic locomotion as a proof-of-concept. The robot is composed of 3 actuated modules connected along the actuated axes with mechanical latching. The three modules are controlled with an external Arduino Uno board using three different signals (Figure 7a). A driving signal controls the contraction and expansion of a module along the actuated axis. At each step cycle, the three modules contract in sequence from left to right (see Fig. 7a) and then expand very rapidly with the same order, thus producing contraction waves used in peristaltic locomotion [24]. This actuation pattern with different speed in contraction and expansion generates a directional friction on the ground which allows the assembly to move forward. The worm's head position has been tracked through motion capture system and the result (Fig. 7b) shows a travel distance of about 15 mm per cycle and a speed of 90cm/min. This speed could be further improved by adding directional patterning to the ventral surface of the robot.

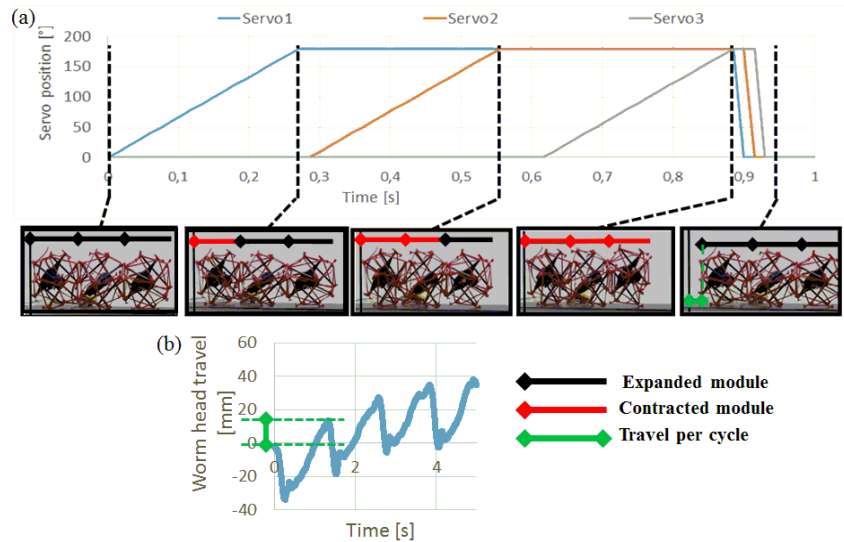

**Fig. 7.** (a) Contraction snapshots sequence of a peristaltic movement cycle compared with the servo position driving signal given by the Arduino board. (b) Graph of the worm's head movement in the longitudinal direction. Every cycle the worm moves forward of about 6.5 mm.

## 5   Conclusions and Future Work

In this paper, we presented a design principle for novel, bio-inspired tensegrity soft modular robots. Although the proposed design displays many desirable features that could lead to the assembly of a variety of more complex robots with a diverse set of behaviors, other tensegrity structure modules could be considered for specialized functions within a heterogeneous modular robot. However, several challenges remain to be addressed. In addition to a connection system that provides more functionalities than simple mechanical latching, as shown in this paper, the system could benefit from a better integrated actuation system to replace the conventional servo-motor used here as proof of concept: a possible solution could be a contraction system made of shape memory alloys that fit the reticular structure of the modules.
Furthermore, autonomous modular robots require at least a sensing system, an internal signaling system, a programmable control unit, and an energy system.

For sensing, we are currently considering replacing the elastic material of the cable network with conductive elastic materials that change conductivity when stretched, thus, enabling a simple form of proprioception. More conventional sensors, such as infrared and vision could be hosted within the cell body to perceive the environment with relatively little interference from the cable and struts.

For internal communication, the tensegrity approach could exploit a phenomenon known as mechanotransduction in biology, which is used by cells to activate biochemical processes or gene expression. This form of communication could enhance or replace digital electrical communication by propagating mechanical disturbances that alter the function and behavior of the cells. Finally, a microcontroller and energy storage could be placed as payloads in the empty volume of the cells (as shown in Figure 2a).

Tensegrity simulation tools, such as the NASA Tensegrity Robotics toolkit [25], could be adapted to design and even evolve [20] modular tensegrity robots.

The scalability of tensegrity structures, from stadium domes to biological cells, opens the possibility of conceiving multi-cellular robots at a smaller scale by means of inkjet printing [26], MEMS fabrication [27], or 3D pop-up micro-structures [28].

We believe that, despite the many challenges that remain to address, the biological inspiration from multi-cellular organisms and the recruitment of novel soft robotic technologies and materials could lead to the generation of diverse, resilient, and scalable modular robots based on tensegrity structures.